\documentclass[10pt,twocolumn,letterpaper]{article}

\usepackage{wacv}
\makeatletter
\@namedef{ver@everyshi.sty}{}
\makeatother
\usepackage{times}
\usepackage{epsfig}
\usepackage{graphicx}
\usepackage{amsmath}
\usepackage{amssymb}
\usepackage{booktabs}
\usepackage{float}
\usepackage[accsupp]{axessibility} 

\def\wacvPaperID{1175} 

\wacvalgorithmstrack   

\wacvfinalcopy 

\ifwacvfinal
\usepackage[breaklinks=true,bookmarks=false]{hyperref}
\else
\usepackage[pagebackref=true,breaklinks=true,colorlinks,bookmarks=false]{hyperref}
\fi

\usepackage[capitalize]{cleveref}
\crefname{section}{Sec.}{Secs.}
\Crefname{section}{Section}{Sections}
\Crefname{table}{Table}{Tables}
\crefname{table}{Tab.}{Tabs.}

\usepackage{multirow}
\usepackage[table,xcdraw]{xcolor}
\usepackage{color, colortbl}
\definecolor{LightCyan}{rgb}{0.88,1,1}
\definecolor{DarkGreen}{rgb}{0,0.7,0}
\usepackage{svg}

\newcommand{\method}{DVF\xspace}
\def\*#1{\mathbf{#1}}

\begin{document}

\title{Dense Voxel Fusion for 3D Object Detection}
\author{
Anas Mahmoud \ \ \ \ Jordan S. K. Hu \ \ \ \ Steven L. Waslander\\
University of Toronto Robotics Institute\\
{\tt\small \{nas.mahmoud, jordan.hu\}@mail.utoronto.ca, steven.waslander@robotics.utias.utoronto.ca }
}
\maketitle
\thispagestyle{empty}

\begin{abstract}
Camera and LiDAR sensor modalities provide complementary appearance and geometric information useful for detecting 3D objects for autonomous vehicle applications. 
However, current end-to-end fusion methods are challenging to train and underperform state-of-the-art LiDAR-only detectors. Sequential fusion methods suffer from a limited number of pixel and point correspondences due to point cloud sparsity, or their performance is strictly capped by the detections of one of the modalities. 
Our proposed solution, Dense Voxel Fusion (DVF) is a sequential fusion method that generates multi-scale dense voxel feature representations, improving expressiveness in low point density regions. To enhance multi-modal learning,  we train directly with projected ground truth 3D bounding box labels, avoiding noisy, detector-specific 2D predictions. Both DVF and the multi-modal training approach can be applied to any voxel-based LiDAR backbone.
DVF ranks $3^{rd}$ among published fusion methods on KITTI's 3D car detection benchmark without introducing additional trainable parameters, nor requiring stereo images or dense depth labels. In addition, DVF significantly improves 3D vehicle detection performance of voxel-based methods on the Waymo Open Dataset.
\end{abstract}
\section{Introduction}
\label{sec:intro}
Both camera and LiDAR sensors are widely used to enable 3D perception tasks for autonomous vehicles. These sensors have differing strengths and weaknesses in terms of range, resolution, and robustness to lighting and weather conditions~\cite{bounding_box_painting}. For instance, LiDAR point clouds provide high quality range information to 3D objects. However, the sparsity of LiDAR returns increases as a function of distance~\cite{MMF}, leading to poorly resolved objects at long range. On the other hand, images from cameras provide a dense representation with fine-grained texture and color information even at significant range, but provide no direct depth measurement of 3D objects.
\begin{figure}[t]
\centering
\includegraphics[width=\columnwidth]{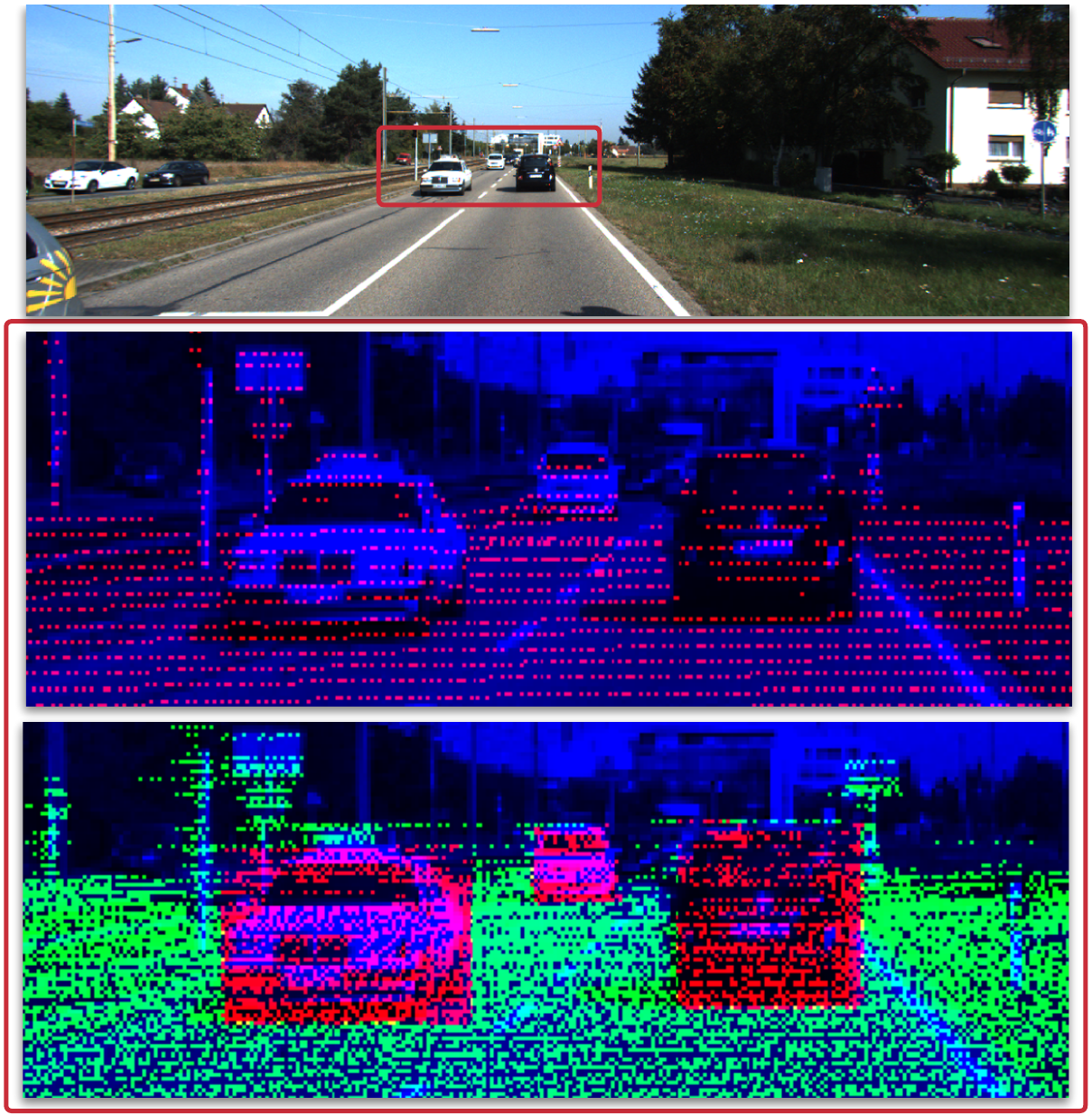}
 \caption{\textbf{Top:} an input image, \textbf{middle:} a cropped, zoomed-in image, where \textcolor{red}{\textbf{red}} points represent LiDAR returns projected onto the image plane, \textbf{bottom:} multi-scale dense voxel centers projected onto the image plane, where \textcolor{DarkGreen}{\textbf{green}} and \textcolor{red}{\textbf{red}} points are associated with background and foreground voxel features, respectively. \method increases the number of correspondences between image and LiDAR features.}
\label{fig:intro}
\end{figure}
Given this complementary information, 3D object detection models using camera and LiDAR should outperform  their LiDAR-only counterparts. However, sequential fusion~\cite{FpointNet, FConvNet, clocs, pointpainting} either suffers from a limited number of pixel and point correspondences due to point cloud sparsity, or their performance is strictly capped by the detections of one of the modalities. Furthermore, end-to-end fusion~\cite{continousFusion, 3D-CVF, EPNet, MMF, AVOD} currently underperforms relative to state-of-the-art LiDAR-only models on the KITTI and Waymo Open Dataset 3D object detection benchmarks~\cite{Kitti, Waymo}. 
One potential reason for this performance gap is that training multi-modal networks is more challenging than uni-modal networks~\cite{MultiModalTrain}. First, multi-modal networks tend to have more trainable parameters and are thus prone to overfitting. In addition, different modality backbones overfit and generalize at different rates, but end-to-end fusion models are usually trained jointly with a single optimization strategy which can lead to sub-optimal solutions~\cite{MultiModalTrain}. Further, the misalignment between the viewpoint of features concatenated from LiDAR bird’s-eye-view (BEV) and image feature maps in current fusion architectures~\cite{AVOD, MV3D} is not ideal for 3D object detection tasks, and leads to reduced performance~\cite{pointpainting}. Sequential fusion methods fuse image predictions either at the input of the LiDAR-only detector~\cite{pointpainting, FpointNet, FConvNet}, or by fusing the output of 3D LiDAR detectors with image predictions~\cite{clocs}. However, current sequential methods have limited fusion capabilities as the recall of 3D objects is strictly capped by the 2D predictions~\cite{FpointNet} or image predictions are fused only at the sparse projected point cloud~\cite{pointpainting}. Finally, effective LiDAR data-augmentation techniques like ground truth sampling~\cite{SECOND}, which accelerate the convergence of LiDAR-only methods, cannot be extended to fusion methods directly because of the missing correspondence of the added LiDAR ground truth objects in the image data.

To resolve the identified issues, we propose Dense Voxel Fusion (\method), a sequential fusion method, that first assigns voxel centers to the 3D location of the occupied LiDAR voxel features. Voxel centers are then projected to the image plane to sample the foreground weight from image-based 2D bounding box predictions. The corresponding voxel features are then fused with image predictions using a parameter-free weighting approach. To enhance multi-modal learning,  we train directly with ground truth projected 3D bounding boxes, avoiding noisy, detector-specific 2D predictions. We also use LiDAR ground truth sampling to both simulate missed 2D detections and to accelerate training convergence. We summarize our approach with two contributions: 
\\[0.2\baselineskip]
\noindent\textbf{Dense Voxel Fusion}. We propose a novel dense voxel fusion method that fuses LiDAR data at the voxel feature representation level with image-based 2D bounding box predictions. Weighting voxel features instead of the sparse LiDAR point features results in dense sampling of image information (see \Cref{fig:intro}), improving the detection of occluded (see~\Cref{tab:painting_car}) and distant objects. To improve robustness against image false detections, we propose a weighting approach that minimizes the coupling of voxel and image features by weighting voxel features using image predictions, and by adding a skip connection to propagate voxel features associated with image missed detections through the LiDAR backbone. This leads to detections that are not strictly capped by image predictions. Finally, \method does not introduce new learnable parameters and can be used with any voxel-based 3D object detector.
\\[0.2\baselineskip]
\noindent \textbf{Multi-Modal Training}. We reason that training using accurate ground truth projected 3D labels while simulating image false detections eliminates the reliance on a specific 2D object detector and generalizes better than training using erroneous 2D object detector predictions. To this end, we propose training \method using ground truth foreground masks generated by 3D bounding box projections. To train robust fusion models, we propose using ground truth sampling~\cite{SECOND} to simulate missed image detections. This encourages \method to learn to detect objects in the LiDAR point cloud that are missed by the image stream and enables using ground truth sampling to accelerate learning convergence. Moreover, since amodal 2D bounding boxes are used, occluded foreground objects are implicitly used to simulate image false positives. During inference time, we propose using pixel-wise foreground aggregation of predicted 2D bounding boxes from any 2D object detector. 

\method outperforms baseline detectors and existing fusion methods when applied to voxel-based detectors. Our proposed training strategy also results in better generalization compared to training using erroneous predictions from 2D object detectors. Our test results on KITTI 3D object detection benchmark~\cite{Kitti} rank $3^{rd}$ among published fusion methods without introducing additional trainable parameters, nor requiring stereo images or dense depth labels. \method also generalizes to the substantially larger Waymo Open Dataset~\cite{Waymo} dataset, improving performance of voxel-based methods on the \textit{val} set.

\begin{figure*}[t!]
  \centering
  \includegraphics[width=1.0\linewidth]{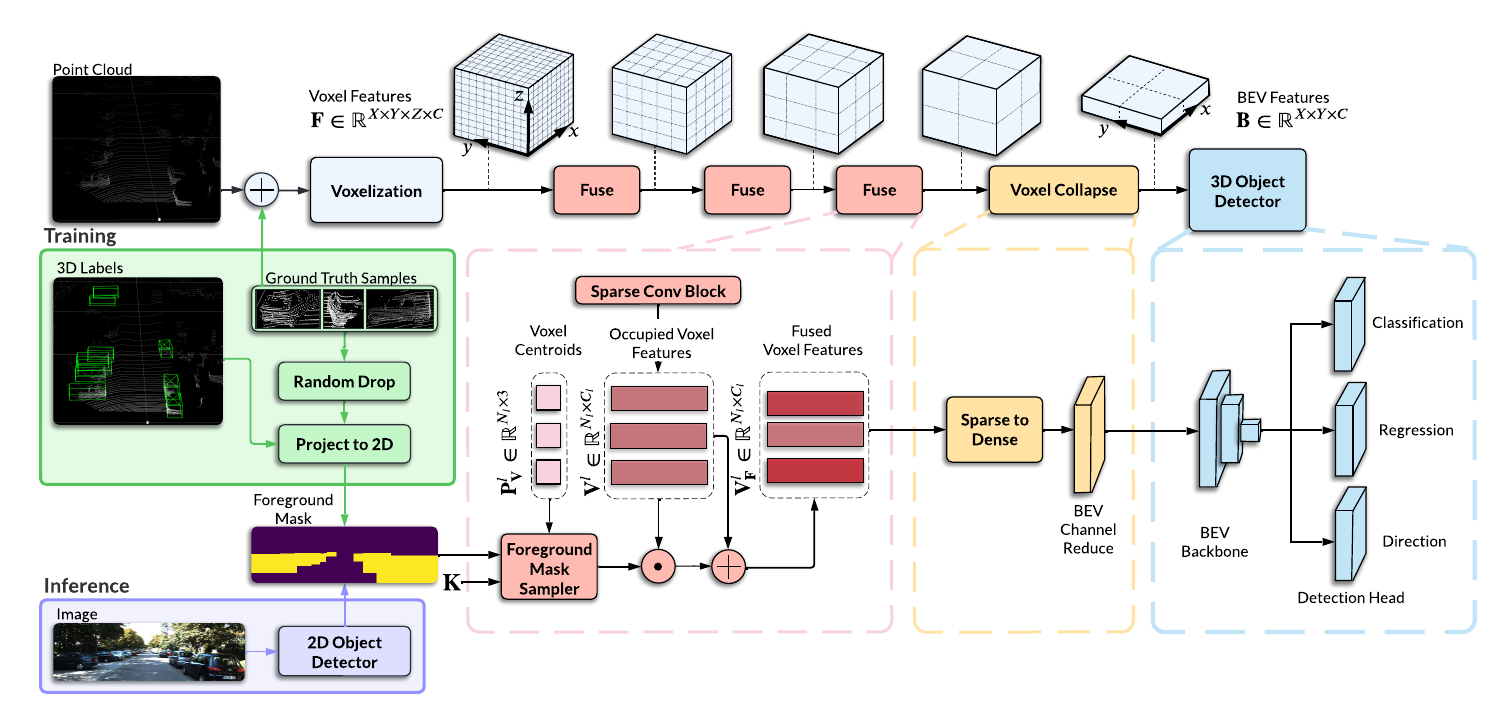}
  \caption{Dense Voxel Fusion (Section~\ref{subsec:overall_framework}) can be applied to any voxel-based LiDAR detector without introducing new learnable parameters. During training (\textcolor{green}{green}), ground truth 3D bounding boxes are projected onto the image plane to construct the 2D foreground mask. In addition, ground truth samples from other scenes are used to simulate image missed detections by randomly dropping corresponding projected 3D bounding boxes from the 2D foreground mask. During inference (\textcolor{violet}{purple}), predicted 2D bounding boxes from any 2D object detector and their associated confidences are used to construct a predicted foreground heatmap. The fusion module (Section~\ref{subsec:dense_fusion}) utilizes centers of multi-scale occupied voxels to localize voxel features and densely sample the foreground heatmap encouraging the propagation of voxel features towards foreground objects. The proposed multi-modal training strategy (Section~\ref{subsec:training_strategy}) retains the performance of the underlying single and two-stage LiDAR detector while utilizing the predicted 2D foreground heatmap to boost the performance of 3D detectors.}
  \label{fig:framework}
\end{figure*}
\section{Related Work}
\label{sec:related_work}
Camera and LiDAR fusion methods can either be trained jointly, in an end-to-end fashion, or independently. We categorize fusion methods based on their training scheme. Here, we present methods that only rely on ground truth 3D or 2D bounding box labels. 
\\[0.2\baselineskip]
\noindent\textbf{End-to-End Fusion}. 
MV3D~\cite{MV3D} and AVOD~\cite{AVOD} utilize an independent backbone for each sensor modality. Features are extracted by projecting a shared set of 3D anchors on BEV and image features. Clearly, fusing misaligned features is not ideal for the task of 3D object detection~\cite{pointpainting}. ContFuse~\cite{continousFusion} and MMF~\cite{MMF} fuse image and LiDAR BEV features using a continuous fusion layer. ContFuse~\cite{continousFusion} attempts to learn a dense BEV representation by querying the nearest $K$ 3D points corresponding to each BEV pixel, and then projecting $K$ points onto the image plane to extract image features. Image features are then used to densify the BEV representation. A drawback of this approach is feature smearing, where multiple image feature vectors are associated with each BEV feature vector. To address the issue of limited correspondences between BEV and image features, MMF~\cite{MMF} learns a pixelwise dense depth map to construct Pseudo-LiDAR points \cite{pseudolidar}, where continuous fusion is applied on the dense point cloud. However, improvement from dense fusion is relatively small and no improvement is observed in 3D AP on the KITTI dataset~\cite{Kitti}. To address the challenge of view misalignment between the LiDAR's BEV and the camera's front view, 3D-CVF~\cite{3D-CVF} proposes combining camera and LiDAR voxel features using the cross-view spatial feature fusion strategy, while EPNet~\cite{EPNet} uses a point-based geometric stream~\cite{PointRCNN}. One of the main challenges of jointly training the LiDAR and image streams is the different overfitting rate of each sensor modality backbone~\cite{MultiModalTrain}. In addition, ground truth sampling~\cite{SECOND}, an effective LiDAR data-augmentation method, is not directly extendable to end-to-end fusion methods due to the difficulty of inserting additional objects into the image stream.

\noindent\textbf{Sequential Fusion}. 
Early sequential fusion methods rely on a two-stage seeding approach. F-PointNet~\cite{FpointNet} and F-ConvNet \cite{FConvNet} reduce the search space for 3D objects, by constraining box predictions to lie within frustums generated from 2D bounding box detections. The main drawback of seeding methods is that the recall of 3D objects is strictly capped by the recall of the 2D object detector. To address this issue, PointPainting~\cite{pointpainting} generates semantic information for each pixel, and projects 3D points into the image to associate semantic information with each point to improve 3D object detection. In addition, FusionPainting~\cite{xu2021fusionpainting} fuses 2D and 3D semantic segmentation predictions and then paints the input point cloud using the fused semantic labels. A drawback of point painting methods is the limited number of correspondences between 3D points and 2D image pixels as seen in the second row of~\Cref{fig:intro}. This is especially problematic for occluded and mid-to-long range objects, where the LiDAR returns are quite sparse. In addition, the painted version of PointRCNN~\cite{PointRCNN} submitted to KITTI~\cite{Kitti} severely degrades the performance of the LiDAR-only baseline~\cite{pointpainting}. We reason using predicted segmentation scores for training is sub-optimal as performance is limited by noisy segmentation predictions. Finally, CLOCs~\cite{clocs} fuses the output predictions of 2D and 3D object detectors by learning to leverage multi-modal geometric and semantic consistencies, in order to predict accurate confidence scores for 3D detections. Since CLOCs is a late fusion method, detections missed by the LiDAR-only detector due to sparsity of LiDAR returns cannot be recovered from image predictions.

\method addresses the aforementioned issues by densely fusing image predictions with LiDAR voxel features while avoiding noisy, detector-specific 2D predictions during training. In addition, \method is capable of using LiDAR ground truth sampling to simulate missed 2D detections and to accelerate training convergence.
\section{Methodology}
\label{sec:metho}

\subsection{Overall Framework}
\label{subsec:overall_framework}
The framework of \method is shown in~\Cref{fig:framework}. The network takes a point cloud as input, in addition to 3D ground truth boxes during training or 2D detections during inference. The ground truth 3D bounding boxes are projected onto the image plane to simulate 2D detections, and pixels within the 2D bounding boxes are assigned a foreground confidence sampled from a uniform distribution $\mathcal{U}_{[a,b]}$. Further, we leverage ground truth sampling to insert additional objects into the LiDAR point cloud, while randomly dropping their corresponding 2D masks in the image plane in order to simulate missed image detections. This training approach only uses LiDAR data and 3D object annotations, making the network training independent of the image-based 2D detector used at inference time (see~\Cref{tab:no_training}).

The point cloud is then voxelized, and a 3D sparse convolution is applied to the voxel grid. Each occupied voxel feature vector is then assigned the voxel center location and subsequently projected onto the image plane to sample from the image foreground mask. The voxel features are then weighted using the sampled 2D foreground confidence. To maintain detection performance despite image false negatives and enable extraction of contextual features from background voxels, a skip connection is added to the weighted voxel features which are passed to the next convolutional block. This process is repeated at the output of each convolutional layer to densely sample the foreground mask. The last convolutional layer features are then projected into a BEV grid and a detection head is used to predict the bounding box proposals. During the testing phase, predicted bounding boxes from any 2D object detector and their corresponding confidence values are used to construct a foreground confidence heatmap. The proposed dense fusion backbone can be used to boost the performance of any voxel-based LiDAR detector while not requiring matched 2D image labels during the training phase.

\begin{figure}[t!]
  \centering
  \includegraphics[scale=0.5]{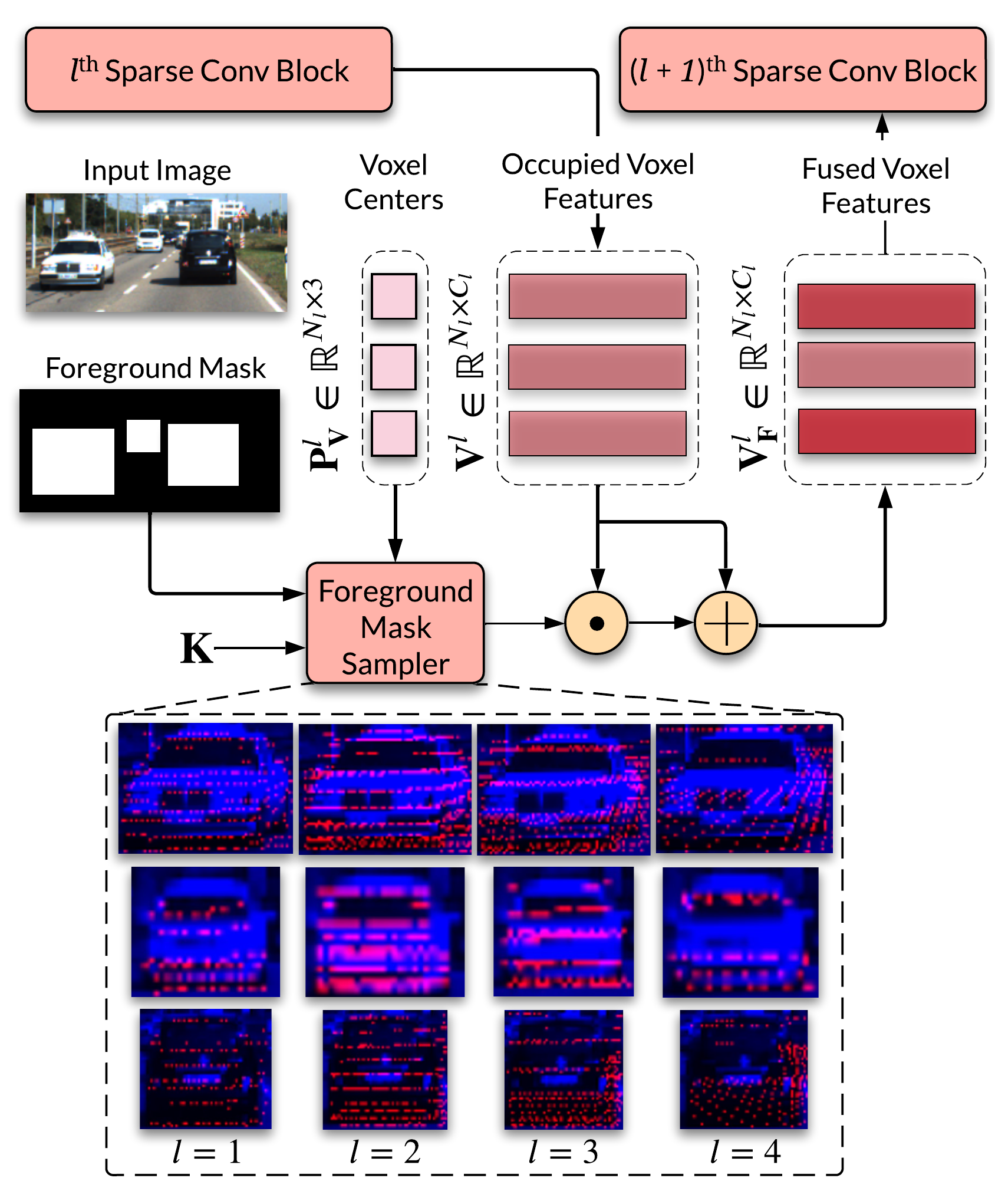}
   \caption{Illustration of Dense Voxel Fusion. The foreground mask is sampled using a set of voxel centers $\mathbf{p}^l_\mathbf{v}$ associated with occupied voxel features $\mathbf{V}^l$ learned by $l^{th}$ sparse convolution block. Fused features from block $l$ are processed by block $l+1$. The projection of the foreground voxels (shown as \textcolor{red}{red} points) in sets $\mathbf{p}^1_\mathbf{v}$, $\mathbf{p}^2_\mathbf{v}$, $\mathbf{p}^3_\mathbf{v}$, $\mathbf{p}^4_\mathbf{v}$ is depicted for the $3$ cars in the scene. At each $l$, a new set of centers is used to sample the mask at new pixel locations, resulting in a dense correspondence between voxel features and image pixels. \Cref{fig:intro} shows voxel and pixel correspondences aggregated from all 4 blocks.} 
   \label{fig:densefusion}
\end{figure}

\subsection{Dense Voxel Fusion}
\label{subsec:dense_fusion}
Due to the occlusion and sparsity inherent in LiDAR point clouds, we propose a dense fusion approach at the voxel level that augments LiDAR point cloud information with dense details from image data. We fuse a foreground mask constructed from the predicted 2D bounding box of any 2D object detector with any voxel-based LiDAR stream. The fusion occurs between the sparse convolution blocks of the Region Proposal Network (RPN). Each block computes a set of voxel features $\mathbf{V}^{l} \in \mathbb{R}^{N_l \times C_l}$, where $l$ is the index of the convolutional block, $N_l$ is the number of occupied voxels and $C_l$ is the number of channels computed by the $l^{th}$ block. Each voxel feature $\mathbf{v}^l_i \in \mathbb{R}^{C_l}$ is assigned a 3D point $\mathbf{p}^l_{\mathbf{v}_i}$. Here, $\mathbf{p}^l_{\mathbf{v}_i}$ is an instance of the set $\mathbf{P}^l_{\mathbf{V}} \in \mathbb{R}^{N_l \times 3}$ and corresponds to the center of the occupied voxel $\mathbf{v}^l_i$. Now that each voxel feature is localized using known camera calibration parameters $\mathbf{K}$, $\mathbf{P}^l_{\mathbf{V}}$ are projected onto the image plane to generate a set of 2D pixel locations $\mathbf{P}^l_\mathbf{C} \in \mathbb{R}^{N_l \times 2}$. Each pixel location $\mathbf{p}^l_{\mathbf{c}_i} \in \mathbb{R}^2$ is used to sample the foreground confidence $\rho_i^l \in [0.0, 1.0]$ via 2D interpolation from the foreground mask.

\begin{figure}[t!]
  \centering
  \includegraphics[scale=0.5]{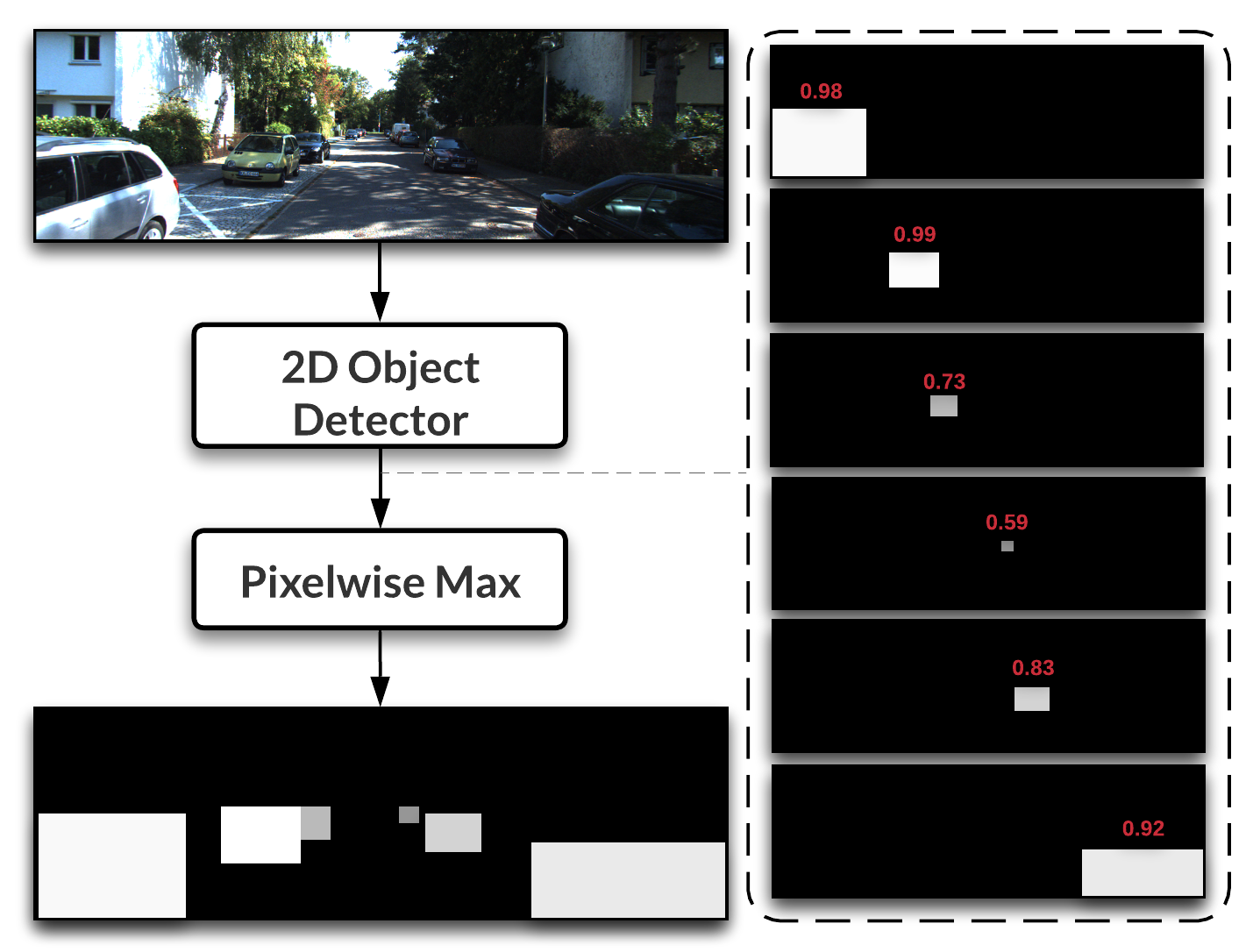}
  \caption{2D foreground heatmap generation during testing time. Darker pixels indicate lower probability of a pixel belonging to a foreground object.}
  \label{fig:predictions}
  \vspace{-8mm}
\end{figure}

To fuse sampled pixel foreground confidence and voxel features, we propose a parameter-free weighting function that minimizes the coupling of learned voxel features and image-based 2D predictions. This is contrary to sparse fusion methods \cite{pointpainting, bounding_box_painting} that concatenate semantic and point features at the input point cloud resulting in the coupling of multi-modal feature extraction, and thus might not be robust to image false positives and false negatives. The proposed method weights each voxel feature $\mathbf{v}_i^l$ with the sampled pixel foreground confidence $\rho_i^l$ using:
 \begin{align}
     \mathbf{v}^l_{\mathbf{f}_i} =  \rho_i^l \mathbf{v}_i^l  + \mathbf{v}_i^l   \quad \forall i \in [0, N_l-1] \label{eq:fusion}
\end{align}
where $\mathbf{v}^l_{\mathbf{f}_i}$ is the weighted feature vector. By weighting voxel features based on the foreground confidence, fused voxel feature vectors stay in the same feature space prior to the fusion step (i.e., $\mathbf{v}^l_i$ , $\mathbf{v_f}_i^l \in  \mathbb{R}^{C_l}$) effectively decoupling geometric feature extraction from erroneous 2D semantic labels. To address the challenge of image false negatives, a skip connection is added to preserve the voxel features where objects are not detected by the camera stream. In addition, by preserving occupied background voxel features, contextual information can be aggregated which is useful for detecting and accurately localizing foreground objects. 

\Cref{fig:densefusion} depicts \method applied to the output of 4 sequential sparse convolutional blocks. The sparse convolutional block consists of a downsampling step which reduces the resolution of the 3D voxel grid by a factor of 2. Due to the downsampling step and the propagation of voxel features to neighbouring voxels after the convolution operation, a new set of voxel features are computed resulting in a new set of voxel centers at the output of each block $l$. To enable dense fusion, multi-scale voxel centers are used to sample the foreground mask. \Cref{fig:densefusion} shows the foreground voxel centers projected onto the 3 foreground objects in the scene (shown as red points). Here, we show the projection of foreground centers at the output of blocks $l = 1, \dots, 4$. Repeatedly sampling the foreground mask at multi-scale voxel centers leads to an increase in the number of correspondences between the voxel features and image-pixels far beyond the sparse correspondence between 3D points and image-pixels.

\subsection{Multi-Modal Training Strategy}
\label{subsec:training_strategy}
Training multi-modal networks is challenging mainly due to the different rates at which the backbone of each modality overfits~\cite{MultiModalTrain}. To overcome this issue, the image-based 2D object detector is not trained jointly with the 3D detector. Therefore, \method is robust to image false positives and false negatives that are otherwise introduced when jointly training a 2D object detector.  Our training strategy consists of training using ground truth projected 3D bounding boxes while simulating image-based false positives and negatives throughout the training phase. Our key finding (see~\Cref{tab:training strategy}) is that training the 3D detector with noise-free ground truth foreground masks while simulating 2D object detection failures generalizes much better on single and two-stage detectors when compared to training with predicted 2D bounding boxes. 

\noindent\textbf{Foreground Heatmap Generation}
As depicted in \Cref{fig:framework}, during the training phase, 3D ground truth bounding boxes are used to construct the foreground mask. The corners of each 3D bounding box are projected onto the image plane using known camera calibration parameters. Then, an axis-aligned 2D bounding box is computed for each set of corners associated with a foreground object in the scene. Since 3D ground truth bounding boxes are used, the proposed fusion method does not require 2D image labels, nor does it require linking of 2D and 3D ground truth boxes. Training directly with ground truth 2D bounding box labels ensures all foreground objects in the scene are accurately recalled by the image stream, providing the fusion model with consistent correspondence between LiDAR and image information. To simulate varying class confidence of predicted bounding boxes, the class confidence of projected 3D bounding boxes is drawn from a uniform distribution $\mathcal{U}_{[a,b]}$. In our experiments, we set $a=0.8$ and $b=1.0$ to simulate detections with high confidence.

\Cref{fig:predictions} illustrates the foreground mask generated during testing time. For input image $\mathbf{I} \in \mathbb{R}^{H \times W \times 3}$, a set of $M$ 2D bounding boxes $\mathbf{B} \in \mathbb{R}^{N \times 4}$ are predicted. $\mathbf{b}^m \in \mathbf{B}$ corresponds to the $m^{th}$ bounding box boundaries denoted by $u^m_1$, $u^m_2$, $v^m_1$ and $v^m_2$. Each $\mathbf{b}^m$ is associated with a predicted confidence $c^m \in [0.0, 1.0]$. \Cref{fig:predictions} shows $M=6$ predictions. For each $\mathbf{b}^m$, a foreground mask $\mathbf{f_r} \in \mathbb{R}^{H \times W}$ is populated using:
\begin{align}
     \mathbf{fr}^{m}(i,j) = \begin{cases}
                           c^m;  & \text{if } v_1 ^m\leq i \leq v_2^m \\
                                &   u_1^m \leq j \leq u_2^m \\
                           0.0;  & \textrm{otherwise}
                          \end{cases}
    \quad \forall i, j \label{eq:b_m_mask}
\end{align}
The foreground mask used for inference aggregates $M$ individual foreground masks using the pixel-wise max operator in~(\ref{eq:pixel-wise}) to preserve high confidence values where foreground objects are  partially occluded by each other (e.g., second and third prediction from top in~\Cref{fig:predictions}).
\begin{align}
     \mathbf{fr}_{agg}(i,j) = \max \{\mathbf{fr}^{0}(i,j), ..., \mathbf{fr}^{M-1}(i,j)\} \quad \forall i, j \label{eq:pixel-wise}
\end{align}

\noindent\textbf{Simulating Missed Detections} To simulate image false negatives during training, $K$ random samples corresponding to ground truth objects from point cloud scenes in the training set are appended to the point cloud. A random subset of the $K$ samples are not projected onto the foreground mask and thus, the 3D object detector is trained to detect foreground objects that are completely missed by the image stream. \method adds a skip connection to the learned voxel features, which preserves the voxel features of foreground objects incorrectly labelled as background voxels due to image missed detections, enabling the use of ground truth sampling.  Training LiDAR detectors with ground truth sampling~\cite{SECOND} accelerates convergence, however, many fusion methods, do not employ ground truth sampling due to the missing image correspondence, which can result in a large gap between fusion and LiDAR-only detectors~\cite{bounding_box_painting}. Our multi-modal training strategy enables utilizing ground truth sampling for accelerating convergence of fusion models while improving robustness against image missed detections. 


\section{Experimental Results}
\label{sec:experi}
We evaluate the effectiveness of \method on KITTI 3D object detection benchmark~\cite{Kitti} and Waymo Open Dataset~\cite{Waymo} applied to: (1) Voxel-based single-stage detector SECOND~\cite{SECOND}, (2) Voxel-based two-stage detector Voxel-RCNN~\cite{voxelrcnn} and (3)  Point-Voxel-based two-stage detector PV-RCNN~\cite{PVRCNN}. The KITTI 3D object detection benchmark~\cite{Kitti} is split into 7,481 and 7,518 training and testing samples respectively. The training samples are further split into 3,712 samples for \textit{train} set and 3,769 samples for \textit{val} set~\cite{kitti_data_split}. Our evaluation is done using average precision with 40 recall points $AP|_{R40}$ and an IoU threshold of 0.7, 0.5 and 0.5 for car, pedestrian and cyclist classes respectively. Waymo Open Dataset~\cite{Waymo} consists of 798 training and 202 validation sequences. We train on 10\% from the training set (15,467 samples) and evaluate on the truncated point cloud within the front camera field-of-view (50.4\textdegree). Waymo Open Dataset uses both standard average precision (AP) and heading (APH) for evaluation metrics with an IoU threshold of 0.7 for vehicles. Waymo Open Dataset is split into two difficulties: LEVEL\_1 only includes 3D labels with more than five LiDAR points, while LEVEL\_2 considers all 3D labels with at least one LiDAR point. 
\\[0.2\baselineskip]
\noindent\textbf{Input Parameters} The initial 3D grid is constructed based on the range of the point cloud input and the grid resolution set for each dimension in 3D space. KITTI uses a point cloud range of $[0.0, 70.0]$, $[-40.0, 40.0]$ and $[-1.0, 3.0]$  meters that are voxelized into a grid resolution of $(0.05, 0.05, 0.1)$ along $X$, $Y$ and $Z$, respectively. Waymo Open Dataset uses a point cloud range of $[0.0, 75.2]$, $[-75.2, 75.2]$ and $[0, 4]$ meters and a grid resolution of $(0.1, 0.1, 0.15)$ along $X$, $Y$ and $Z$, respectively. Only points in the field-of-view of the front camera are considered in both datasets.
\begin{table}[t]
\centering
\resizebox{\columnwidth}{!}{
\begin{tabular}{c|c|c|ccc}
\toprule
\multirow{2}{*}{Method} & \multirow{2}{*}{Modality} & \multirow{2}{*}{Reference} & \multicolumn{3}{c}{Car - 3D AP}                  \\
                        &                           &                            & Easy           & Mod.           & Hard           \\ \hline
SA-SSD~\cite{SA-SSD}                  & L                         & CVPR-2020                  & 88.75          & 79.79          & 74.16          \\
PV-RCNN~\cite{PVRCNN}                 & L                         & CVPR-2020                  & 90.25          & 81.43          & 76.82          \\
Voxel R-CNN~\cite{voxelrcnn}             & L                         & AAAI-2021                  & 90.90           & 81.62          & 77.06          \\
CT3D~\cite{CT3D}             & L                         & ICCV-2021                  & 87.83           & 81.77          & 77.16          \\
Pyramid-PV~\cite{pyramid_rcnn}              & L                         & ICCV-2021                  & 88.39          & 82.08          & 77.49          \\
VoTr-PV~\cite{VTor}                 & L                         & ICCV-2021                  & 89.90          & 82.09          & \textbf{79.14}          \\
SPG-PV~\cite{SPG}                  & L                         & ICCV-2021                  & 90.50           & 82.13          & 78.90           \\
SE-SSD~\cite{se_ssd}                  & L                         & CVPR-2021                  & \textbf{91.49}          & 82.54          & 77.15          \\ 
BtcDet~\cite{BTC}                & L & AAAI-2022 & 90.64 & \textbf{82.86}	& 78.09 \\ \hline \hline
MV3D~\cite{MV3D}                    & L$+$I                      & CVPR-2017                  & 74.97          & 63.63          & 54.00          \\
ContFuse~\cite{continousFusion}                & L$+$I                       & ECCV-2018                  & 83.68          & 68.78          & 61.67          \\
F-PointNet~\cite{FpointNet}              & L$+$I                       & CVPR-2018                  & 82.19          & 69.79          & 60.59          \\
AVOD-FPN~\cite{AVOD}                & L$+$I                        & IROS-2018                  & 83.07          & 71.76          & 65.73          \\
MMF~\cite{MMF}                     & L$+$I                        & CVPR-2019                  & 88.40           & 77.43          & 70.22          \\
EPNet~\cite{EPNet}                   & L$+$I                       & ECCV-2020                  & 89.81          & 79.28          & 74.59          \\
3D-CVF~\cite{3D-CVF}                  & L$+$I                       & ECCV-2020                  & 89.20          & 80.05          & 73.11          \\
CLOCs-PV~\cite{clocs}                & L$+$I                       & IROS-2020                  & 88.94          & 80.67          & 77.15          \\ \hline
\textbf{DVF-PV (ours)}  &L$+$I                       & -                          & \textbf{90.99} & \textbf{82.40} & \textbf{77.37} \\ \bottomrule
\end{tabular}}
\caption{3D detection results on the KITTI \textit{test} set using $AP|_{R_{40}}$ metric. Models ending with PV use PV-RCNN as their base LiDAR architecture.  L and I represent the LiDAR point cloud and the camera image, respectively.} 
\label{tab:test}
\vspace{-4mm}
\end{table}
\\[0.2\baselineskip]
\noindent\textbf{Training and Inference Details} The sparse convolution backbone with \method is implemented in PyTorch~\cite{pytorch_dl}. All baselines and \method~models are trained using 2 NVIDIA Tesla P100 GPUs with a batch size of $2$ for $80$ epochs on KITTI and 4 NVIDIA Tesla V100 GPUs with a batch size of $4$ for $50$ epochs on Waymo Open Dataaset. Adam~\cite{adam} optimizer and one-cycle learning rate policy is used~\cite{one_cycleLR}. 

During training, LiDAR-based global data-augmentation strategies are applied, including global scaling, global rotation around $Z$ axis, and random flipping along the $X$ axis. For \method models, all global linear transformations of the point cloud are constructed and then reversed before using known camera parameters to sample either the ground truth foreground mask or the predicted foreground heatmap. For training \method models, we set the number of ground truth samples $K$ to 5 for each class and the drop-out percentage to 50$\%$.  During training phase, the foreground mask is constructed using 3D ground truth bounding box projections. For inference, pixel-wise aggregation, as outlined in (\ref{eq:pixel-wise}), is used to compute the input predicted foreground heatmap using of Cascade-RCNN~\cite{crcnn} 2D detections. For experiments on KITTI \textit{val} set, we use 2D detections from Cascade-RCNN~\cite{crcnn} published by CLOCs~\cite{clocs}. For KITTI \textit{test} set results, we train Cascade-RCNN implemented in Detectron2~\cite{detectron2} on \textit{train} and \textit{val} set. For inference on Waymo Open Dataset \textit{val} set, 2D predictions from a Cascade-RCNN~\cite{crcnn} detector pre-trained on COCO dataset is used.

\subsection{KITTI Dataset Results}
\Cref{tab:test} shows the results of \method applied to PV-RCNN~\cite{PVRCNN} (\method-PV) on the KITTI~\cite{Kitti} \textit{test} set compared to state-of-the-art LiDAR-only and Camera and LiDAR fusion methods. Each group is listed in order of performance of the moderate difficulty level car class. DVF-PV outperforms state-of-the-art fusion method CLOCs~\cite{clocs} on $AP|_{R_{40}}$ by +2.05\%, +1.73\% and +0.22\% on the car class on easy, moderate, and hard difficulties, respectively. It is important to note that DVF-PV and CLOCs-PV both use PV-RCNN~\cite{PVRCNN} and Cascade-RCNN~\cite{crcnn} for 3D and 2D detections respectively. Moreover, DVF-PV outperforms PV-RCNN~\cite{PVRCNN} on $AP|_{R_{40}}$ by +0.74\%, +0.97\% and +0.55\% on the car class on easy, moderate, and hard difficulties, respectively. PV-RCNN~\cite{PVRCNN} learns foreground segmentation of voxel features using LiDAR point supervision yet underperforms compared to weighting voxel features using dense image predictions via DVF.  SE-SSD~\cite{se_ssd} is a LiDAR single stage detector that employs a consistency loss between teacher and student predictions.  While BtcDet~\cite{BTC} learns object shape priors for partially occluded objects. These techniques are orthogonal to our proposed fusion framework and incorporating them will be a goal for future work. Finally, current state-of-the-art fusion method SFD~\cite{sfd} requires dense depth labels to train the depth completion network, while VPFNet~\cite{vpfnet} requires stereo images. Both methods have only been tested on KITTI dataset and can only be applied to two-stage detectors. 

We also show in \Cref{tab:baseline_car} that \method can be used to improve the performance of Voxel-based single-stage detectors  SECOND~\cite{SECOND}, Voxel-based two-stage detector Voxel-RCNN~\cite{voxelrcnn} and Point-Voxel-based two-stage detectors PV-RCNN~\cite{PVRCNN}. The relative gains achieved on the \textit{val} set for \method + PV-RCNN generalize to the \textit{test} set. On the contrary, using the same 3D and 2D detectors, CLOCs-PV~\cite{clocs} achieves comparable performance to \method on \textit{val} set, though these gains do not persist on the \textit{test} set.

\subsection{Waymo Open Dataset Results} 
\Cref{tab:waymo} shows the results of \method on the larger and more complex Waymo Open Dataset \textit{val} set. Even with the lower voxel resolution, \method provides meaningful dense voxel features with an increase of +1\% AP/APH on SECOND, a +2\% AP/APH increase on Voxel-RCNN and +5\% AP/APH increase on PV-RCNN. We attribute the large gain on Waymo Open Dataset to the intra-class variance within the Waymo Open Dataset vehicle class. Unlike KITTI, Waymo Open Dataset vehicles vary in size and local geometry significantly, encapsulating both small objects, such as motorcycles, and large objects, such as trucks. The 2D foreground image information provides complementary information to gauge different vehicle types for accurate 3D object detection, especially when leveraging a second stage for bounding box refinement.

\begin{table}[t]
\centering
\resizebox{\columnwidth}{!}{
\begin{tabular}{c|ccc|ccc}
\toprule
                                         & \multicolumn{3}{c|}{Car - 3D AP}                                                                                         & \multicolumn{3}{c}{Car - BEV AP}                 \\
\multirow{-2}{*}{Method}                 & Easy                                   & Mod.                                   & Hard                                   & Easy           & Mod.           & Hard           \\ \hline
SECOND                                   & 90.29                                  & 81.83                                  & 79.01                                  & 92.60           & 90.02          & 88.05          \\
\method + SECOND                             & \textbf{92.03}                         & \textbf{82.84}                         & \textbf{79.72}                         & \textbf{94.43} & \textbf{90.85} & \textbf{88.35} \\
\rowcolor{LightCyan} 
\textit{Improvement}                     & \textit{+1.74}                          & \textit{+1.01}                          & \textit{+0.71}                          & \textit{+1.83}  & \textit{+0.83}  & \textit{+0.30}   \\ \hline
\cellcolor[HTML]{FFFFFF}Voxel-RCNN       & \cellcolor[HTML]{FFFFFF}92.44          & \cellcolor[HTML]{FFFFFF}85.20           & \cellcolor[HTML]{FFFFFF}82.89          & 95.58          & 91.30           & 89.01          \\
\cellcolor[HTML]{FFFFFF}\method + Voxel-RCNN & \cellcolor[HTML]{FFFFFF}\textbf{92.96} & \cellcolor[HTML]{FFFFFF}\textbf{85.81} & \cellcolor[HTML]{FFFFFF}\textbf{83.20}  & \textbf{96.24} & \textbf{91.93} & \textbf{89.55} \\
\rowcolor{LightCyan}
\textit{Improvement}                     & \textit{+0.52}                          & \textit{+0.61}                          & \textit{+0.31}                          & \textit{+0.66}  & \textit{+0.63}  & \textit{+0.54}  \\ \hline
\cellcolor[HTML]{FFFFFF}PV-RCNN          & \cellcolor[HTML]{FFFFFF}91.88          & \cellcolor[HTML]{FFFFFF}84.83          & \cellcolor[HTML]{FFFFFF}82.55          & 93.73          & 90.65          & 88.55          \\
\cellcolor[HTML]{FFFFFF}\method + PV-RCNN     & \cellcolor[HTML]{FFFFFF}\textbf{93.07} & \cellcolor[HTML]{FFFFFF}\textbf{85.84} & \cellcolor[HTML]{FFFFFF}\textbf{83.13} & \textbf{96.21} & \textbf{91.66} & \textbf{89.17} \\
\rowcolor{LightCyan} 
\textit{Improvement}                     & \textit{+1.19}                          & \textit{+1.01}                          & \textit{+0.58}                          & \textit{+2.48}  & \textit{+1.01}  & \textit{+0.62}  \\ \bottomrule
\end{tabular}}
\caption{3D detection results on KITTI \textit{val} set for SECOND~\cite{SECOND}, Voxel-RCNN~\cite{voxelrcnn} and PV-RCNN~\cite{PVRCNN} on car class. Results are shown using 3D and BEV AP $|_{R_{40}}$.}
\label{tab:baseline_car}
\end{table}
\begin{table}[]
\centering
  \label{tab:waymo}
\resizebox{\columnwidth}{!}{
\begin{tabular}{c|cc|cc}
\toprule
                         & \multicolumn{2}{c|}{Vehicle LEVEL\_1} & \multicolumn{2}{c}{Vehicle LEVEL\_2} \\
\multirow{-2}{*}{Method} & AP              & APH            & AP             & APH            \\ \hline
SECOND                   & 60.93           & 60.35          & 56.43          & 55.89          \\
SECOND + DVF             & \textbf{61.99}  & \textbf{61.41} & \textbf{57.45} & \textbf{56.91} \\
\rowcolor[HTML]{E0FFFF} 
\textit{Improvement}     & \textit{+1.06}  & \textit{+1.06} & \textit{+1.02} & \textit{+1.02} \\ \hline
Voxel-RCNN               & 64.96           & 64.32          & 60.15          & 59.55          \\
Voxel-RCNN + DVF         & \textbf{67.17}  & \textbf{66.53} & \textbf{62.18} & \textbf{61.58} \\
\rowcolor[HTML]{E0FFFF} 
\textit{Improvement}     & \textit{+2.21}  & \textit{+2.21} & \textit{+2.03} & \textit{+2.03} \\ \hline
PV-RCNN                  & 62.32           & 61.30          & 57.65          & 56.70          \\
PV-RCNN + DVF            & \textbf{67.62}  & \textbf{67.09} & \textbf{62.66} & \textbf{62.17} \\
\rowcolor[HTML]{E0FFFF} 
\textit{Improvement}     & \textit{+5.30}  & \textit{+5.79} & \textit{+5.01} & \textit{+5.47} \\ \bottomrule
\end{tabular}}
 \caption{3D detection results on Waymo Open Dataset \textit{val} set for SECOND~\cite{SECOND}, Voxel-RCNN~\cite{voxelrcnn}, and PV-RCNN~\cite{PVRCNN}. Results are shown using AP and APH.}
\end{table}
\begin{table}
\centering
\resizebox{\columnwidth}{!}{
\begin{tabular}{c|ccc|ccc}
\toprule
                         & \multicolumn{3}{c|}{\begin{tabular}[c]{@{}c@{}}\textbf{C-RCNN}\\  Car - 3D AP\end{tabular}} & \multicolumn{3}{c}{\begin{tabular}[c]{@{}c@{}}\textbf{GT}\\  Car - 3D AP\end{tabular}} \\
\multirow{-2}{*}{Method} & Easy                            & Mod.                            & Hard                           & Easy                            & Mod.                           & Hard                           \\ \hline
Painted SECOND     & 91.26                           & 82.27                          & 79.43   & 91.24                           & 82.31                           & 79.57                                                    \\
\method + SECOND        & \textbf{92.03}                  & \textbf{82.84}                 & \textbf{79.72}      & \textbf{92.11}                  & \textbf{85.36}                  & \textbf{82.64}                                  \\
\rowcolor{LightCyan}
\textit{Improvement}   & \textit{+0.77}                   & \textit{+0.57}                  & \textit{+0.29}  & \textit{+0.87}                   & \textit{+3.05}                   & \textit{+3.07}                                    \\ \hline
\rowcolor[HTML]{FFFFFF} 
Painted Voxel-RCNN  & 92.74                           & 84.08                          & 82.54  & 92.73                           & 84.99                           & 83.11                                                    \\
\rowcolor[HTML]{FFFFFF} 
\method + Voxel-RCNN        &\textbf{92.96}                  & \textbf{85.81}                 & \textbf{83.20}  & \textbf{95.50}                   & \textbf{86.72}                  & \textbf{85.70}                                    \\
\rowcolor{LightCyan}
\textit{Improvement}  & \textit{+0.22}                   & \textit{+1.73}                  & \textit{+0.66}    & \textit{+2.77}                   & \textit{+1.73}                   & \textit{+2.59}                                   \\ \hline
\rowcolor[HTML]{FFFFFF} 
Painted PV-RCNN   & 92.47                           & 85.22                          & 82.68   & 92.85                           & 86.20                            & 83.66                                                    \\
\rowcolor[HTML]{FFFFFF} 
\method + PV-RCNN           & \textbf{93.07}                  & \textbf{85.84}                 & \textbf{83.13}   & \textbf{94.44}                  & \textbf{86.71}                  & \textbf{86.22}                                  \\
\rowcolor{LightCyan}
\textit{Improvement}   & \textit{+0.60}                    & \textit{+0.62}                  & \textit{+0.45}  & \textit{+1.59}                   & \textit{+0.51}                   & \textit{+2.56}                                    \\ \bottomrule
\end{tabular}}
\caption{Comparison between Pointpainting~\cite{pointpainting} and \method. The same training strategy is used for both fusion models. Inference on the KITTI \textit{val} set is performed using both using Cascade-RCNN~\cite{crcnn} 2D predictions \textbf{C-RCNN} and projected ground truth 3D bounding boxes \textbf{GT}. Improvements in 3D$AP|_{R_{40}}$ are shown relative to Painted models.}
\label{tab:painting_car}
\end{table}

\subsection{Ablation Studies}

We provide ablation studies on our proposed fusion method and training strategy.
\\[0.2\baselineskip]
\noindent\textbf{Point versus Voxel Fusion}
In ~\Cref{tab:painting_car}, we compare the performance of \method to PointPainting~\cite{pointpainting}, a sparse fusion method. Both fusion methods use the proposed training strategy of using ground truth 2D labels while simulating image false detections. We conduct inference using 2D predictions from Cascade-RCNN~\cite{crcnn}. In addition, during inference, we utilize ground truth 2D bounding boxes for all models to determine an upper bound on the potential gains for PointPainting and \method. Using Cascade-RCNN predictions, \method achieves a relative 3D $AP|_{R_{40}}$ gain over PointPainting of +0.57\%, +1.73\%, +0.62\% for SECOND~\cite{SECOND}, Voxel-RCNN~\cite{voxelrcnn} and PV-RCNN~\cite{PVRCNN} for car moderate settings, respectively. This can be attributed to \method substantially increasing the number of correspondences between image pixels and multi-scale occupied voxel centers compared to the limited number of correspondences between image pixels and 3D points.  In addition, \method achieves a consistent relative gain over PointPainting of more than +2.50\% on hard setting for all models when ground truth 2D labels $\mathbf{GT}$ are used, indicating that densely fusing features is especially beneficial for far-away and occluded objects with fewer LiDAR points. Experiments with $\mathbf{GT}$ 2D labels demonstrate the significant gain of \method models over sparse methods with improvements in 2D object detectors. Finally, we also show in~\Cref{tab:multi-class} that \method can generalize to a multi-class setting.
\begin{table}[t]
\centering
\resizebox{\columnwidth}{!}{
\begin{tabular}{c|ccc|ccc|ccc}
\toprule
                                             & \multicolumn{3}{c|}{Car - 3D AP} & \multicolumn{3}{c|}{Pedestrian - 3D AP} & \multicolumn{3}{c}{Cyclist - 3D AP} \\
\multirow{-2}{*}{Method}                     & Easy      & Mod.      & Hard     & Easy        & Mod.        & Hard        & Easy       & Mod.       & Hard      \\ \hline
Painted SECOND                               & \textbf{90.87}     & 81.16     & 78.79    & \textbf{57.65}       & \textbf{53.38}       & \textbf{49.16}       & 82.34      & 65.49      & 61.77     \\
\cellcolor[HTML]{FFFFFF}\method+SECOND           & 90.83     & \textbf{82.12}     & \textbf{79.17}    & 57.00       & 52.82       & 48.70       & \textbf{86.72}      & \textbf{66.70}      & \textbf{62.68}     \\
\rowcolor{LightCyan}\textit{Improvement} & \textit{-0.04}     & \textit{+0.96}      & \textit{+0.38}     & \textit{-0.65}       & \textit{-0.56}       & \textit{-0.47}       & \textit{+4.37}       & \textit{+1.21}       & \textit{+0.91}      \\ \hline
Painted VoxelRCNN                            & 92.38     & 84.42     & 82.71    & 68.56       & 62.09       & 57.34       & 90.08      & 71.74      & \textbf{68.27}     \\
\cellcolor[HTML]{FFFFFF}\method+VoxelRCNN        & \textbf{92.45}     & \textbf{85.28}     & \textbf{82.84}    & \textbf{70.13}       & \textbf{62.76}       & \textbf{57.65}       & \textbf{90.93}      & \textbf{72.60}      & 68.24     \\
\rowcolor{LightCyan}\textit{Improvement} & \textit{+0.07}      & \textit{+0.85}      & \textit{+0.13}     & \textit{+1.57}        & \textit{+0.67}        & \textit{+0.31}        & \textit{+0.85}       & \textit{+0.86}       & \textit{-0.03}     \\ \hline
Painted PV-RCNN                               & 92.30     & 84.85     & 82.75    & 65.32       & 58.20       & 53.82       & 88.95      & 70.87      & 67.33     \\
\cellcolor[HTML]{FFFFFF}\method+PV-RCNN           & \textbf{92.34}     & \textbf{85.25}     & \textbf{82.97}    & \textbf{66.08}       & \textbf{59.18}       & \textbf{54.68}       & \textbf{90.93}      & \textbf{72.46}      & \textbf{68.05}     \\
\rowcolor{LightCyan}\textit{Improvement} & \textit{+0.04}      & \textit{+0.40}      & \textit{+0.22}     & \textit{+0.75}       & \textit{+0.98}        & \textit{+0.86}        & \textit{+1.98}       & \textit{+1.59}       & \textit{+0.72}      \\ \bottomrule
\end{tabular}}
\caption{Multi-class 3D detection results on KITTI \textit{val} set. Improvements in 3D$AP|_{R_{40}}$ are shown relative to Painted models.}
\label{tab:multi-class}
\end{table}
\\[0.2\baselineskip]
\noindent\textbf{Effect of Point Cloud Sparsity}
\method increases the number of correspondences between LiDAR and image data as seen in~\Cref{fig:intro}. This is contrary to Pointpainting~\cite{pointpainting} where fusion is at the 3D point level, and is therefore sparse, especially at mid-to-long range. We conduct an experiment on the effect of point cloud sparsity on the relative performance between Pointpainting and \method using SECOND~\cite{SECOND}. In~\Cref{tab:sparsity_experiment} we randomly drop 10\% and 20\% of the points and train SECOND models. \method outperforms Pointpainting over all classes. We attribute this to densely fusing image detections at the voxel rather than the point level. This also shows that \method can be useful when LiDAR returns become more sparse in adverse weather conditions (i.e., rain or snow).
\begin{table}[t]
\centering
\resizebox{\columnwidth}{!}{
\begin{tabular}{cc|ccc|ccc|ccc}
\toprule
\multicolumn{1}{c|}{\multirow{2}{*}{Method}} & \multirow{2}{*}{\begin{tabular}[c]{@{}c@{}}Points \\ Dropped\end{tabular}} & \multicolumn{3}{c|}{Car - 3D AP} & \multicolumn{3}{c|}{Pedestrian - 3D AP} & \multicolumn{3}{c}{Cyclist - 3D AP} \\
\multicolumn{1}{c|}{}                        &                                                                            & Easy      & Mod.      & Hard     & Easy        & Mod.        & Hard        & Easy       & Mod.       & Hard      \\ \hline
\multicolumn{1}{c|}{Painted SECOND}        & \multirow{2}{*}{10\%}                                                      & \textbf{90.06}     & 80.03     & 77.22    & 56.38       & 51.09       & 46.64       & 80.28      & 62.51      & 58.71     \\
\multicolumn{1}{c|}{DVF + SECOND}            &                                                                            & 89.58     & \textbf{81.74}     & \textbf{78.93}    & \textbf{57.33}       & \textbf{52.71}       & \textbf{48.32}       & \textbf{84.11}      & \textbf{63.97}      & \textbf{60.30}     \\ \hline
\rowcolor{LightCyan}\multicolumn{2}{c|}{Improvement}                                                                                          & \textit{-0.48}     & \textit{+1.71}     & \textit{+1.70}    & \textit{+0.95}       & \textit{+1.62}       & \textit{+1.68}       & \textit{+3.83}      & \textit{+1.46}      & \textit{+1.59}     \\ \hline
\multicolumn{1}{c|}{Painted SECOND}        & \multirow{2}{*}{20\%}                                                      & 90.11     & 79.36     & 76.72    & 55.65       & 50.69       & 45.99       & 80.66      & 62.02      & 58.16     \\
\multicolumn{1}{c|}{DVF + SECOND}            &                                                                            & \textbf{90.51}     & \textbf{81.54}     & \textbf{78.18}    & \textbf{57.32}       & \textbf{52.61}       & \textbf{47.89}       & \textbf{82.51}      & \textbf{62.31}      & \textbf{58.76}     \\ \hline
\rowcolor{LightCyan}\multicolumn{2}{c|}{Improvement}                                                                                          & \textit{+0.40}     & \textit{+2.18}     & \textit{+1.46}    & \textit{+1.67}       & \textit{+1.92}       & \textit{+1.90}       & \textit{+1.86}      & \textit{+0.28}      & \textit{+0.61}     \\ \bottomrule
\end{tabular}}
\caption{Effect of point cloud sparsity on DVF and pointpainting. The points dropped column corresponds to the percentage of point cloud dropped before training both models.}
\label{tab:sparsity_experiment}
\vspace{-5mm}
\end{table}
\begin{table}[]
\centering
\resizebox{\columnwidth}{!}{
\begin{tabular}{cccccc}
\toprule
\multicolumn{1}{c|}{}                                                     & \multicolumn{2}{c|}{2D Source}                                                        & \multicolumn{3}{c}{Car - 3D AP}                                                                                                                           \\
\multicolumn{1}{c|}{\multirow{-2}{*}{Method}}                             & C-RCNN                                       & \multicolumn{1}{c|}{GT}                        & Easy                                              & Mod.                                              & Hard                                              \\ \hline
\multicolumn{1}{c|}{}                                                     & \checkmark                                            & \multicolumn{1}{c|}{}                          & 90.37                                             & 81.39                                             & 78.69                                             \\
\multicolumn{1}{c|}{\multirow{-2}{*}{\method + SECOND}}                             & \multicolumn{1}{l}{}                         & \multicolumn{1}{c|}{\checkmark}                         & \multicolumn{1}{l}{\textbf{92.03}}                         & \multicolumn{1}{l}{\textbf{82.84}}                         & \multicolumn{1}{l}{\textbf{79.72}}                         \\ \hline
\rowcolor[HTML]{FFFFFF} 
\rowcolor{LightCyan}\multicolumn{3}{c|}{Improvement}                                                                                                           & \textit{+1.66}                                     & \textit{+1.45}                                      & \textit{+1.03}                                     \\ \hline
\rowcolor[HTML]{FFFFFF} 
\multicolumn{1}{c|}{\cellcolor[HTML]{FFFFFF}}                             & \checkmark                                            & \multicolumn{1}{c|}{\cellcolor[HTML]{FFFFFF}}  & 92.42                                             & 85.14                                             & 82.86                                             \\
\rowcolor[HTML]{FFFFFF} 
\multicolumn{1}{c|}{\multirow{-2}{*}{\cellcolor[HTML]{FFFFFF}\method + Voxel-RCNN}} & \multicolumn{1}{l}{\cellcolor[HTML]{FFFFFF}} & \multicolumn{1}{c|}{\cellcolor[HTML]{FFFFFF}\checkmark} & \multicolumn{1}{l}{\cellcolor[HTML]{FFFFFF}\textbf{92.96}} & \multicolumn{1}{l}{\cellcolor[HTML]{FFFFFF}\textbf{85.81}} & \multicolumn{1}{l}{\cellcolor[HTML]{FFFFFF}\textbf{83.20}}  \\ \hline
\rowcolor[HTML]{FFFFFF} 
\rowcolor{LightCyan}\multicolumn{3}{c|}{Improvement}                                                                                                         & \textit{+0.54}                                     & \textit{+0.67}                                     & \textit{+0.34}                                     \\ \hline
\rowcolor[HTML]{FFFFFF} 
\multicolumn{1}{c|}{\cellcolor[HTML]{FFFFFF}}                             & \checkmark                                            & \multicolumn{1}{c|}{\cellcolor[HTML]{FFFFFF}}  & 92.10                                             & 83.99                                             & 82.73                                             \\
\rowcolor[HTML]{FFFFFF} 
\multicolumn{1}{c|}{\multirow{-2}{*}{\cellcolor[HTML]{FFFFFF}\method + PV-RCNN}}    & \multicolumn{1}{l}{\cellcolor[HTML]{FFFFFF}} & \multicolumn{1}{c|}{\cellcolor[HTML]{FFFFFF}\checkmark} & \multicolumn{1}{l}{\cellcolor[HTML]{FFFFFF}\textbf{93.07}} & \multicolumn{1}{l}{\cellcolor[HTML]{FFFFFF}\textbf{85.84}} & \multicolumn{1}{l}{\cellcolor[HTML]{FFFFFF}\textbf{83.13}} \\ \hline
\rowcolor[HTML]{FFFFFF} 
\rowcolor{LightCyan}\multicolumn{3}{c|}{Improvement}                                                                                                         & \textit{+0.97}                                     & \textit{+1.85}                                     & \textit{+0.40} \\ \bottomrule
\end{tabular}}
\caption{Comparison between training \method using ground truth 2D bounding boxes and Cascade-RCNN~\cite{crcnn} 2D predictions. Inference on KITTI \textit{val} set is performed using Cascade-RCNN 2D predictions.}
\label{tab:training strategy}
\vspace{-5mm}
\end{table}
\\[0.2\baselineskip]
\noindent\textbf{Effect of Training Using Ground Truth Labels}
~\Cref{tab:training strategy} shows the results of training \method models using ground truth 2D labels versus training using 2D predictions from Cascade-RCNN~\cite{crcnn}. Inference is done using predictions from Cascade-RCNN. Training with ground truth labels generalizes better on the \textit{val} set and outperforms training on Cascade-RCNN~\cite{crcnn} predictions by $+1.45\%$, $+0.67\%$, $+1.85\%$ for SECOND~\cite{SECOND}, Voxel-RCNN~\cite{voxelrcnn} and PV-RCNN~\cite{PVRCNN} respectively for the car moderate setting. Training using erroneous bounding box predictions provided by Cascade-RCNN, result in an incorrect association between image predictions and LiDAR voxel features. In addition, adding ground truth-sampled LiDAR objects without projection to the image plane increases the number of 2D missed detections. This leads to a fusion model that learns not to rely on the inconsistent 2D predictions. On the other hand, not using ground truth sampling severely degrades the performance of the fusion model, making it less robust to 2D missed detections. Training directly with ground truth 2D bounding box labels ensures all foreground objects in the scene are accurately recalled by the image stream, providing the fusion model with a consistent correspondence between LiDAR and image information. Since perfect 2D recall is achieved during training, ground truth sampling can be introduced to simulate image missed detections and to accelerate convergence. This multi-modality training approach leads to better generalization compared to erroneous 2D predictions and decouples the training of the image-based 2D detector from the 3D LiDAR detector.

\section{Conclusion}
\label{sec:concl}
We present \method, a novel fusion method that generates multi-scale multi-modal dense voxel feature representations, improving expressiveness in low point density regions.  \method can be applied to any voxel-based LiDAR backbone without introducing additional learnable parameters. Our test results on KITTI 3D object detection benchmark~\cite{Kitti} rank $3^{rd}$ among published fusion methods without introducing additional trainable parameters, nor requiring stereo images or dense depth labels. \method also generalizes to the substantially larger Waymo Open Dataset~\cite{Waymo} dataset, improving performance of voxel-based methods on the \textit{val} set. Our multi-modal training strategy relies solely on LiDAR data and object annotations, making the network training independent of the 2D detector that is used at inference time. 
\clearpage
{\small
\bibliographystyle{ieee_fullname}
\bibliography{main}

\begin{thebibliography}{10}\itemsep=-1pt

\bibitem{crcnn}
Zhaowei Cai and Nuno Vasconcelos.
\newblock Cascade r-cnn: High quality object detection and instance
  segmentation, 2019.

\bibitem{kitti_data_split}
Xiaozhi Chen, Kaustav Kundu, Yukun Zhu, Andrew~G Berneshawi, Huimin Ma, Sanja
  Fidler, and Raquel Urtasun.
\newblock 3d object proposals for accurate object class detection.
\newblock In {\em NeurIPS}, volume~28, 2015.

\bibitem{MV3D}
Xiaozhi Chen, Huimin Ma, Ji Wan, Bo Li, and Tian Xia.
\newblock Multi-view 3d object detection network for autonomous driving.
\newblock In {\em CVPR}, July 2017.

\bibitem{voxelrcnn}
Jiajun Deng, Shaoshuai Shi, Peiwei Li, Wengang Zhou, Yanyong Zhang, and
  Houqiang Li.
\newblock Voxel r-cnn: Towards high performance voxel-based 3d object
  detection.
\newblock {\em AAAI}, pages 1201--1209, 2021.

\bibitem{Kitti}
Andreas Geiger, Philip Lenz, and Raquel Urtasun.
\newblock Are we ready for autonomous driving? the kitti vision benchmark
  suite.
\newblock In {\em CVPR}, 2012.

\bibitem{SA-SSD}
Chenhang He, Hui Zeng, Jianqiang Huang, Xian-Sheng Hua, and Lei Zhang.
\newblock Structure aware single-stage 3d object detection from point cloud.
\newblock In {\em CVPR}, June 2020.

\bibitem{EPNet}
Tengteng Huang, Zhe Liu, Xiwu Chen, and X. Bai.
\newblock Epnet: Enhancing point features with image semantics for 3d object
  detection.
\newblock In {\em ECCV}, 2020.

\bibitem{adam}
Diederik~P. Kingma and Jimmy Ba.
\newblock Adam: A method for stochastic optimization.
\newblock In {\em ICLR}, 2015.

\bibitem{AVOD}
J. {Ku}, M. {Mozifian}, J. {Lee}, A. {Harakeh}, and S.~L. {Waslander}.
\newblock Joint 3d proposal generation and object detection from view
  aggregation.
\newblock In {\em IROS}, 2018.

\bibitem{MMF}
M. {Liang}, B. {Yang}, Y. {Chen}, R. {Hu}, and R. {Urtasun}.
\newblock Multi-task multi-sensor fusion for 3d object detection.
\newblock In {\em CVPR}, 2019.

\bibitem{bounding_box_painting}
Anas Mahmoud and Steven~L. Waslander.
\newblock Sequential fusion via bounding box and motion pointpainting for 3d
  objection detection.
\newblock In {\em CRV}, 2021.

\bibitem{pyramid_rcnn}
Jiageng Mao, Minzhe Niu, Haoyue Bai, Xiaodan Liang, Hang Xu, and Chunjing Xu.
\newblock Pyramid r-cnn: Towards better performance and adaptability for 3d
  object detection.
\newblock In {\em ICCV}, October 2021.

\bibitem{VTor}
Jiageng Mao, Yujing Xue, Minzhe Niu, Haoyue Bai, Jiashi Feng, Xiaodan Liang,
  Hang Xu, and Chunjing Xu.
\newblock Voxel transformer for 3d object detection.
\newblock In {\em ICCV}, October 2021.

\bibitem{clocs}
Su Pang, Daniel Morris, and Hayder Radha.
\newblock Clocs: Camera-lidar object candidates fusion for 3d object detection.
\newblock In {\em IROS}, 2020.

\bibitem{pytorch_dl}
Adam Paszke, Sam Gross, Francisco Massa, Adam Lerer, James Bradbury, Gregory
  Chanan, Trevor Killeen, Zeming Lin, Natalia Gimelshein, Luca Antiga, Alban
  Desmaison, Andreas Kopf, Edward Yang, Zachary DeVito, Martin Raison, Alykhan
  Tejani, Sasank Chilamkurthy, Benoit Steiner, Lu Fang, Junjie Bai, and Soumith
  Chintala.
\newblock Pytorch: An imperative style, high-performance deep learning library.
\newblock In {\em Advances in Neural Information Processing Systems},
  volume~32, 2019.

\bibitem{FpointNet}
Charles~R. Qi, Wei Liu, Chenxia Wu, Hao Su, and Leonidas~J. Guibas.
\newblock Frustum pointnets for 3d object detection from rgb-d data.
\newblock In {\em CVPR}, June 2018.

\bibitem{CT3D}
Hualian Sheng, Sijia Cai, Yuan Liu, Bing Deng, Jianqiang Huang, Xian-Sheng Hua,
  and Min-Jian Zhao.
\newblock Improving 3d object detection with channel-wise transformer.
\newblock In {\em ICCV}, October 2021.

\bibitem{PVRCNN}
Shaoshuai Shi, Chaoxu Guo, L. Jiang, Zhe Wang, Jianping Shi, Xiaogang Wang, and
  Hongsheng Li.
\newblock Pv-rcnn: Point-voxel feature set abstraction for 3d object detection.
\newblock {\em CVPR}, pages 10526--10535, 2020.

\bibitem{PointRCNN}
S. {Shi}, X. {Wang}, and H. {Li}.
\newblock Pointrcnn: 3d object proposal generation and detection from point
  cloud.
\newblock In {\em CVPR}, 2019.

\bibitem{one_cycleLR}
Leslie~N. Smith.
\newblock A disciplined approach to neural network hyper-parameters: Part 1 --
  learning rate, batch size, momentum, and weight decay, 2018.

\bibitem{Waymo}
Pei Sun, Henrik Kretzschmar, Xerxes Dotiwalla, Aurelien Chouard, Vijaysai
  Patnaik, P. Tsui, J. Guo, Y. Zhou, Y. Chai, Benjamin Caine, V. Vasudevan, Wei
  Han, J. Ngiam, Hang Zhao, A. Timofeev, S. Ettinger, Maxim Krivokon, A. Gao,
  Aditya Joshi, Y. Zhang, Jon Shlens, Zhi-Feng Chen, and Dragomir Anguelov.
\newblock Scalability in perception for autonomous driving: Waymo open dataset.
\newblock In {\em CVPR}, 2020.

\bibitem{pointpainting}
Sourabh Vora, Alex~H. Lang, Bassam Helou, and Oscar Beijbom.
\newblock Pointpainting: Sequential fusion for 3d object detection.
\newblock In {\em CVPR}, June 2020.

\bibitem{MultiModalTrain}
Weiyao Wang, Du Tran, and Matt Feiszli.
\newblock What makes training multi-modal classification networks hard?
\newblock In {\em CVPR}, June 2020.

\bibitem{pseudolidar}
Yan Wang, Wei-Lun Chao, Divyansh Garg, Bharath Hariharan, Mark Campbell, and
  Kilian~Q. Weinberger.
\newblock Pseudo-lidar from visual depth estimation: Bridging the gap in 3d
  object detection for autonomous driving.
\newblock In {\em CVPR}, June 2019.

\bibitem{FConvNet}
Z. {Wang} and K. {Jia}.
\newblock Frustum convnet: Sliding frustums to aggregate local point-wise
  features for amodal 3d object detection.
\newblock In {\em IROS}, 2019.

\bibitem{continousFusion}
Zining Wang, Wei Zhan, and Masayoshi Tomizuka.
\newblock Fusing bird view {LIDAR} point cloud and front view camera image for
  deep object detection.
\newblock {\em CoRR}, abs/1711.06703, 2017.

\bibitem{sfd}
Xiaopei Wu, Liang Peng, Honghui Yang, Liang Xie, Chenxi Huang, Chengqi Deng,
  Haifeng Liu, and Deng Cai.
\newblock Sparse fuse dense: Towards high quality 3d detection with depth
  completion.
\newblock In {\em Proceedings of the IEEE/CVF Conference on Computer Vision and
  Pattern Recognition}, pages 5418--5427, 2022.

\bibitem{detectron2}
Yuxin Wu, Alexander Kirillov, Francisco Massa, Wan-Yen Lo, and Ross Girshick.
\newblock Detectron2.
\newblock \url{https://github.com/facebookresearch/detectron2}, 2019.

\bibitem{BTC}
Qiangeng Xu, Yiqi Zhong, and Ulrich Neumann.
\newblock Behind the curtain: Learning occluded shapes for 3d object detection.
\newblock In {\em Proceedings of the AAAI Conference on Artificial
  Intelligence}, volume~36, pages 2893--2901, 2022.

\bibitem{SPG}
Qiangeng Xu, Yin Zhou, Weiyue Wang, Charles~R. Qi, and Dragomir Anguelov.
\newblock Spg: Unsupervised domain adaptation for 3d object detection via
  semantic point generation.
\newblock In {\em ICCV}, October 2021.

\bibitem{xu2021fusionpainting}
Shaoqing Xu, Dingfu Zhou, Jin Fang, Junbo Yin, Zhou Bin, and Liangjun Zhang.
\newblock Fusionpainting: Multimodal fusion with adaptive attention for 3d
  object detection.
\newblock {\em ITSC}, 2021.

\bibitem{SECOND}
Yan Yan, Yuxing Mao, and Bo Li.
\newblock Second: Sparsely embedded convolutional detection.
\newblock {\em Sensors}, 18(10), 2018.

\bibitem{3D-CVF}
Jin~Hyeok Yoo, Yecheol Kim, Jisong Kim, and Jun~Won Choi.
\newblock 3d-cvf: Generating joint camera and lidar features using cross-view
  spatial feature fusion for 3d object detection.
\newblock In {\em ECCV}, 2020.

\bibitem{se_ssd}
Wu Zheng, Weiliang Tang, Li Jiang, and Chi-Wing Fu.
\newblock Se-ssd: Self-ensembling single-stage object detector from point
  cloud.
\newblock In {\em CVPR}, June 2021.

\bibitem{vpfnet}
Hanqi Zhu, Jiajun Deng, Yu Zhang, Jianmin Ji, Qiuyu Mao, Houqiang Li, and
  Yanyong Zhang.
\newblock Vpfnet: Improving 3d object detection with virtual point based lidar
  and stereo data fusion.
\newblock {\em arXiv preprint arXiv:2111.14382}, 2021.

\end{thebibliography}
}

\end{document}